\documentclass{ecai}
\usepackage{latexsym}

\usepackage{graphicx}
\usepackage{amsmath,amsfonts}
\usepackage{algorithm}
\usepackage{bm}
\usepackage{subfigure}
\usepackage{caption}
\usepackage{makecell}
\usepackage{multirow}
\usepackage{diagbox}
\usepackage{booktabs}
\usepackage{adjustbox}
\usepackage{makecell}
\usepackage{setspace}

\usepackage[noend]{algpseudocode}
\usepackage{color}

\algrenewcommand\algorithmicindent{1.0em}

\newcommand{\cfs}{{\cal F}_s}
\newcommand{\cfd}{{\cal F}_d}

\newcommand{\cc}{{\cal C}}
\newcommand{\cd}{{\cal D}}

\algnewcommand{\LeftComment}[1]{\State \(\triangleright\) #1}
\newcommand{\pluseq}{\mathrel{+}=}

\definecolor{cGreen}{RGB}{0,150,0}
\definecolor{brown}{RGB}{139,64,0}
\usepackage{ulem}

\begin{document}

\begin{frontmatter}

    \title{Multi-Domain Learning From Insufficient Annotations}

    \author[A,B]{\fnms{Rui}~\snm{He}\orcid{0000-0002-6338-1387}}
    \author[A]{\fnms{Shengcai}~\snm{Liu}\orcid{0000-0002-4223-2438}\thanks{Corresponding Author. Email: liusc3@sustech.edu.cn}}
    \author[A,C]{\fnms{Jiahao}~\snm{Wu}}
    \author[B]{\fnms{Shan}~\snm{He}}
    \author[A]{\fnms{Ke}~\snm{Tang}}

    \address[A]{Department of Computer Science and Engineering, Southern University of Science and Technology}
    \address[B]{School of Computer Science, University of Birmingham}
    \address[C]{Department of Computing, The Hong Kong Polytechnic University}

    \begin{abstract}
        Multi-domain learning (MDL) refers to simultaneously constructing a model or a set of models on datasets collected from different domains.
        Conventional approaches emphasize domain-shared information extraction and domain-private information preservation, following the shared-private framework (SP models), which offers significant advantages over single-domain learning.
        However, the limited availability of annotated data in each domain considerably hinders the effectiveness of conventional supervised MDL approaches in real-world applications.
        In this paper, we introduce a novel method called multi-domain contrastive learning (MDCL) to alleviate the impact of insufficient annotations by capturing both semantic and structural information from both labeled and unlabeled data.
        Specifically, MDCL comprises two modules: inter-domain semantic alignment and intra-domain contrast.
        The former aims to align annotated instances of the same semantic category from distinct domains within a shared hidden space, while the latter focuses on learning a cluster structure of unlabeled instances in a private hidden space for each domain.
        MDCL is readily compatible with many SP models, requiring no additional model parameters and allowing for end-to-end training.
        Experimental results across five textual and image multi-domain datasets demonstrate that MDCL brings noticeable improvement over various SP models.
        Furthermore, MDCL can further be employed in multi-domain active learning (MDAL) to achieve a superior initialization, eventually leading to better overall performance.
    \end{abstract}

\end{frontmatter}

\section{Introduction}

In many machine learning tasks, models are built on datasets collected from various data sources with different distributions, known as domains.
For instance, texts from different sources like news articles, social media posts, and scientific papers constitute distinct domains in natural language processing.
In computer vision, images of differing styles, such as sketches, cartoons, art paintings, and camera photos \cite{PACS} are considered distinct domains.
While each domain possesses unique information, they often share a significant amount of information with other domains.
Naive solutions involve jointly building a single model across domains or independently constructing models for each domain, as is done in conventional single domain learning (SDL) approaches.
However, joint training may neglect the unique information on each domain, while independent training disregards the correlations among domains \cite{ASP-MTL}.
To address these shortcomings, multi-domain learning (MDL)\cite{mdl} has been proposed to simultaneously capture domain-shared and domain-private information.
Most existing MDL works concentrate on sharing information among domains while preserving domain-private information through models under the shared-private framework (SP models) \cite{ASP-MTL, MAN}.
Typically, following the concept of domain adaptation (DA) \cite{TL-survey}, shared information can be captured through distribution alignment across domains, allowing several DA methods to be utilized in MDL.
Besides, private information is usually managed by a private component of the model for each domain.
Accounting for both types of information has led to significant performance improvements over joint and independent training in the past decade \cite{ASP-MTL}.

In real-world applications, obtaining a sufficiently labeled dataset can be costly, even within a single domain \cite{AL-Comparative-Survey,FSPAP,GACPP}.
This issue exacerbated in MDL since constructing labeled multi-domain dataset is even more challenging due to the difficulty in accessing data from multiple domain experts \cite{mdl-medical-image,MDDA,MDNMT}.
For instance, in the case of multi-domain medical image datasets \cite{mdl-medical-image}, the high cost of manual annotations from medical experts across various research fields is just one of the challenges.
The varying privacy and legal concerns, quality assurance processes, and labeling tools across domains also entail additional costs.
The aforementioned MDL approaches face challenges in this high-cost scenario as they heavily rely on fully supervised training from a relatively sufficiently annotated multi-domain dataset.
Therefore, a natural question arises: \textbf{can we perform cost-efficient MDL with insufficient annotated data?}


To the best of our knowledge, only a few works address the issue of insufficient annotations from multiple domains.
Some works utilize contrastive \cite{prototype-unsup,CLDA} and semi-supervised \cite{ECACL} learning to alleviate the impact of insufficient annotations across domains.
The key is to utilize unlabeled data to improve the performance.
However, these works focus on the domain adaptation problem, where only the target domain performance is concerned and the domain-private information is usually neglected.
Other works propose to utilize active learning (AL) \cite{active-learning-survey} in MDL, which is referred to as multi-domain active learning (MDAL) \cite{mdal}.
Given a budget for annotation, MDAL begins with a small set of labeled instances and iteratively selects the new instances for model building.
However, without a good initial model trained on insufficient labeled instances, the selection process is likely to be biased and unreliable in the subsequent MDAL iterations.
In summary, no method is readily applicable for MDL with insufficient annotations so far.

\begin{figure}[htbp]
    \hspace{-0.05\linewidth}
    \includegraphics[width=1.1\linewidth]{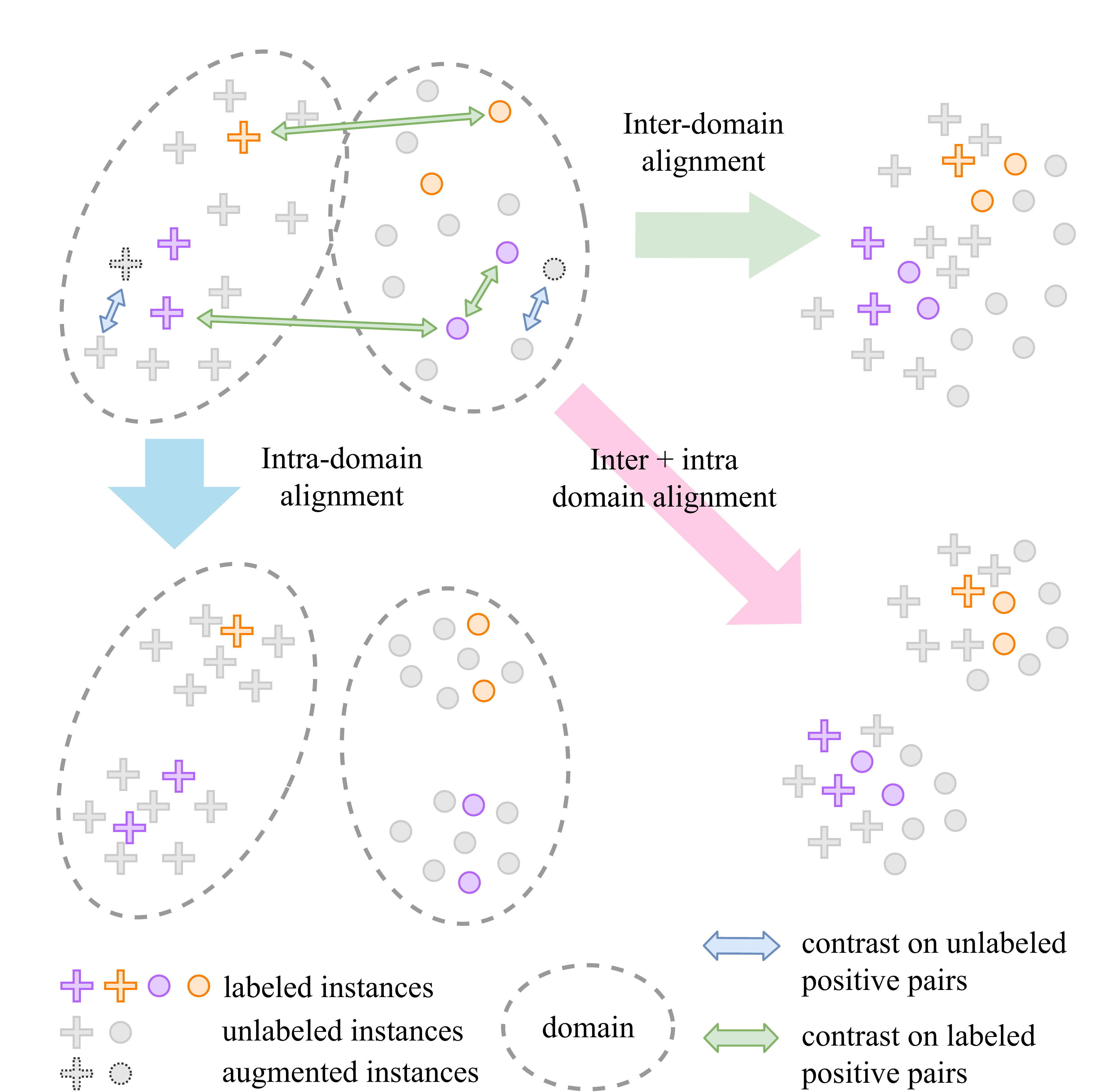}
    \vspace{0.04\linewidth}
    \caption{\textbf{Intuitive understanding of MDCL.}
        An illustrative example in the hidden space.
        (1) Inter-domain alignment aims to align items within the same category but from different domains closer to each other.
        (2) Intra-domain contrast aims to maintain a cluster structure in each domain and make instances more separable.
    }
    \label{fig:intuition}
\end{figure}

In this paper, we propose a novel MDL approach, called multi-domain contrastive learning (MDCL), to construct neural network models on a limited number of labeled instances from each domain.
Figure~\ref{fig:intuition} presents an intuitive understanding of MDCL, where the semantic and structural information are respectively captured from labeled and unlabeled data.
Specifically, MDCL comprises two components: a supervised contrastive loss to align instances of the same category from different domains within a shared hidden space and an unsupervised contrastive loss that focuses on learning the cluster structure of instances from the same domain in a private hidden space.
By integrating both components, MDCL can learn a well-aligned representation from insufficient annotations.
Importantly, MDCL is readily compatible with various share-private (SP) models for MDL \cite{DSN,ASP-MTL,MAN}, requiring no additional model parameters and allowing for end-to-end training.
Experimental results across five textual and image multi-domain datasets demonstrate that MDCL brings noticeable improvement over various SP models.

The main contributions of this paper are summarized as follows:
\begin{itemize}
    \item
          We introduce a novel approach called MDCL to build neural network models on insufficient labels in MDL.
          MDCL is readily compatible with many SP models, requiring no additional model parameters and allowing for end-to-end training.
    \item
          Experimental results demonstrate that MDCL yields noticeable improvement over many SP models across five textual and image multi-domain datasets.
    \item
          Moreover, MDCL can be employed in MDAL to achieve a superior initialization and consequently result a better performance.
\end{itemize}

\begin{figure*}[ht]
    \hspace{0.05\linewidth}
    \subfigure[Share-private backbone (MAN example)]{
        \includegraphics[height=0.4\linewidth]{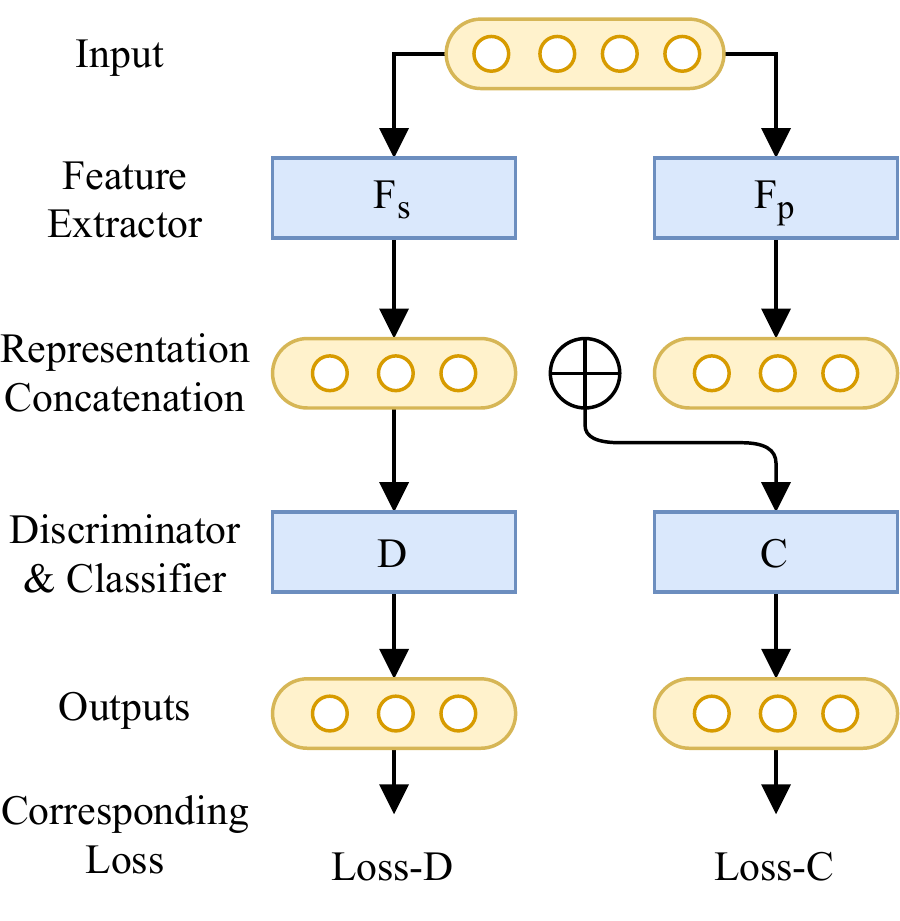}
        \label{fig:man}
    }
    \hspace{0.05\linewidth}
    \subfigure[MDCL components]{
        \includegraphics[height=0.4\linewidth]{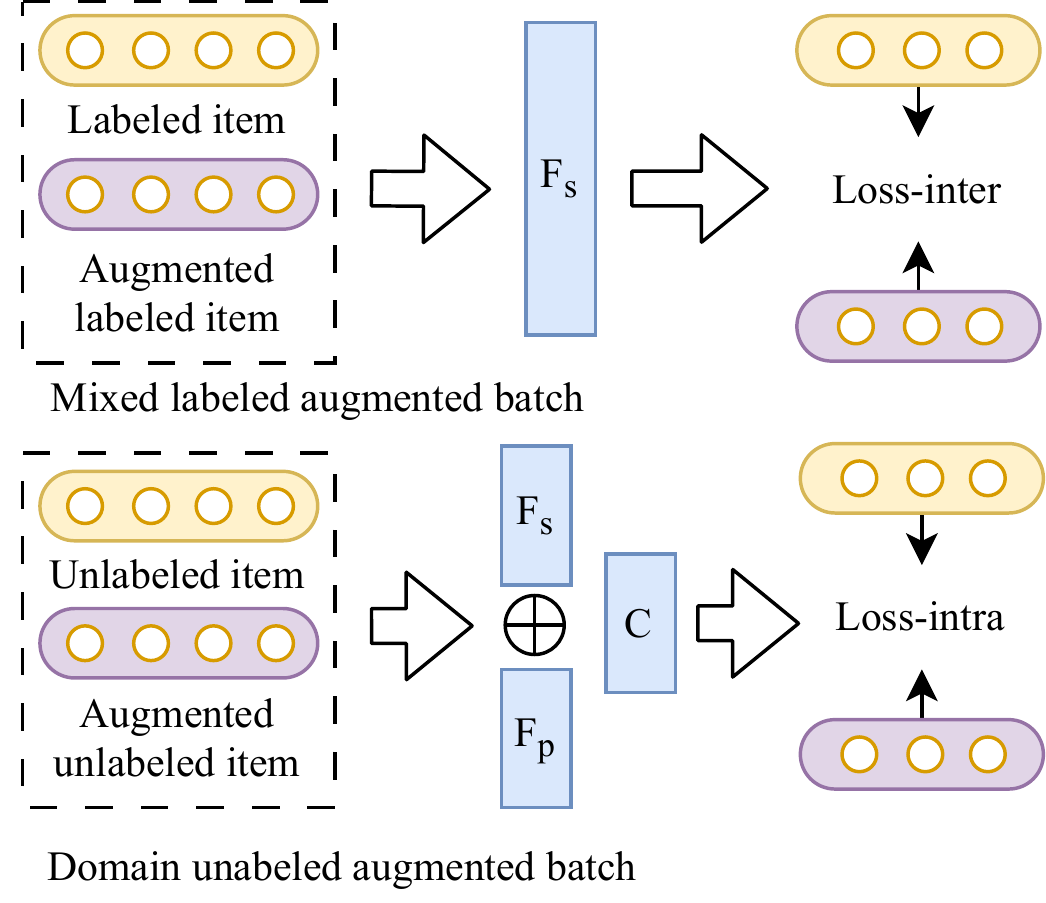}
        \label{fig:mdcl-component}
    }
    \caption{\textbf{Overview of MDCL method.}
        (a) As a representative model for MDL under the share-private framework, MAN is taken as an illustration.
        (b) MDCL comprises two components: an inter-domain semantic alignment and an intra-domain representation learning.
    }
    \label{fig:overview}
\end{figure*}

\section{Related Work}

This work is closely related to multi-domain learning (MDL) \cite{mdl} and contrastive learning \cite{infonce}, which will be briefly reviewed in the following sections.
The representative works and solutions would be included.

\subsection{Multi-Domain Learning}
\label{sec:MDL}

MDL \cite{mdl} differs from both multi-task learning (MTL) \cite{mtl-survey}, domain adaptation (DA) \cite{TL-survey} and domain generalization \cite{DG,LEDG,lu2023large} in that MDL focuses on performing the same task across all domains simultaneously.
By leveraging shared information from multiple related domains, MDL enhances the model's performance on each individual domain.

\paragraph*{Domain-shared information learning}
The domain-shared information learning is a crucial step in MDL, which aims to extract common features across different domains.
This can be achieved by using a shared feature extractor that aligns the marginal distributions of different domains.
The alignment is typically performed by minimizing the maximum mean discrepancy (MMD) between domains \cite{TCA}.
Adversarial training can also be used to handle distribution alignment for more than two domains.
The extracted features are effective if a discriminator cannot distinguish the domain of the instances.
Several models based on this intuition from domain adaptation (DA) can be adapted for use in MDL.
Examples include Domain-adversarial neural networks (DANN) \cite{DANN} and Adversarial discriminative domain adaptation (ADDA) \cite{ADDA}.
Additionally, Feng \textit{et al.} \cite{feng2019self} proposed a method that directly learns representations for MDL using this distribution alignment approach.

\paragraph*{Domain-private information imparted}
In MDL, the domain-invariant representations may not fully capture the unique information of each domain.
To address this issue, several methods have been proposed to capture both domain-shared and domain-private information.
One approach is concatenating the domain-private information with the domain-shared information for making predictions.
For example, Domain separation networks (DSN) \cite{DSN} use separate feature extractors for domain-private and domain-shared information in a DA setting.
Adversarial shared-private model (ASP-MTL) \cite{ASP-MTL} and Multinomial adversarial networks \cite{MAN} apply this idea to MDL.
This architecture that concatenates representations through domain-shared and domain-private feature extractors is commonly known as the \textbf{share-private framework} (SP models).
Empirical study \cite{mdal} has shown the superiority of the share-private framework.

\paragraph*{Multi-domain active learning}

To address the labeling cost in MDL, MDAL \cite{MDAL-text} iteratively selects informative instances to reduce the labeling burden.
A comprehensive comparative study \cite{mdal} has shown that AL can improve performance compared to passive models.
The MAN model under the share-private framework outperforms other methods in both passive and active scenarios.
Moreover, taking into account the evaluation of multi-domain informativeness, perturbation-based two-stage multi-domain active learning (P2S-MDAL) \cite{he2023perturbation} serves as the first ad hoc active learning strategy for MDAL.
Some works have also applied AL to multiple domains but without considering information-sharing schemes.
They either build classifiers on each domain \cite{mdal-AD} or use a single model for all domains \cite{msal}.

\subsection{Contrastive Learning}

\paragraph*{Contrastive learning on single domain}
Contrastive learning (CL) is one of the most prevalent paradigms of self-supervised learning.
The general learning paradigm of CL always performs a contrasting among different views augmented from the original data~\cite{infonce,DCL,supervised-CL}.
Due to the capability of extracting supervision signals from unlabeled data, CL is explored in a wide range of fields and achieves promising performance~\cite{infonce,DCL,contraCoding}, including multi-domain learning.

\paragraph*{Contrastive learning across domains}
Contrastive learning has been applied in DA, where the learning process is guided by a fully labeled source domain to improve performance on the unlabeled or sparsely labeled target domain.
Self-supervised tasks can be directly used to jointly learn a shared feature extractor \cite{unsupervised-DA}.
Besides, class prototypes \cite{prototype-unsup,CLDA,ECACL} can be generated to align the categorical distributions between the source and target domains.
Meanwhile, instance contrastive alignment \cite{CLDA,multilevelCL} can also be used to build a feature extractor without using labels on the target domain to learn the target domain representation.

These studies are most close to our work technically.
However, they only focus on domain adaptation (DA) setting and only employ a single feature extractor for both domains.
They do not consider the MDL problem, where both shared and private information should be taken into account and further been handled by certain model components.
Besides, there is no fully labeled source domain in our setting to extract reliable class prototypes as in \cite{prototype-unsup,CLDA,ECACL}.

\section{Problem Formulation}

Following the definition given in \cite{mdal}, we re-formulate the problem of multi-domain learning, which further considers the insufficient annotations and utilizes both labeled and unlabeled instances.
The formulation is written as follows:

In multi–domain learning, there are $K$ different data sources (domains) $\mathcal{D} = \{\mathcal{D}_1, $ $ \mathcal{D}_2, \dots, \mathcal{D}_K\}$.
A set of data pools $\mathcal{P} = \{\mathcal{P}_1, \mathcal{P}_2,\dots, \mathcal{P}_K\}$ containing both labeled and unlabeled data is collected from $\mathcal{D}$ in advance.
The labeled data from each pool constitute a labeled data set $\mathcal{L} = \{\mathcal{L}_1, $ $ \mathcal{L}_2, \dots, \mathcal{L}_K\}$.
Considering the scenario with limited labeled instances, the number of the labeled instances is much smaller than the collected data pool, i.e. $|\mathcal{L}| \ll |\mathcal{P}|$.
In this scenario, MDL is to find a set of models $\mathcal{M} = \{\mathcal{M}_1,\mathcal{M}_2,\dots,\mathcal{M}_K\}$ for $K$ domains by capturing common knowledge on the unlabeled data pools $\mathcal{P}$ and limited labeled instances $\mathcal{L}$ of different domains, which can be expressed as follows:
\begin{equation}
    \begin{aligned}
        \min_{\mathcal{M}} {Loss}_{\rm sup}(\mathcal{M}; \mathcal{L}) + \Omega (\mathcal{M}; \mathcal{P}, \mathcal{L})
    \end{aligned}
\end{equation}
${Loss}_{\rm sup}(\mathcal{M}; \mathcal{L})$ denotes the supervised loss on the labeled set $\mathcal{L}$.
$\Omega (\mathcal{M}; \mathcal{P}, \mathcal{L})$ denotes a designed loss on the set of data pools $\mathcal{P}$ and limited labeled instances $\mathcal{L}$ for capturing the common knowledge through model $\mathcal{M}$.

\section{Methodology}

This section introduces our novel multi-domain contrastive learning (MDCL) method for MDL with limited annotations.
The key is to maximally utilize the unlabeled data pools and limited labeled instances to learn a set of models for $K$ domains.
The labeled data can be used to capture the semantic information and align the distributions across domains.
The unlabeled data can further be used to learn the structural information of the data distribution and preserve the local information.
MDCL introduces an inter-domain semantic alignment loss and an intra-domain representation learning loss to utilize the unlabeled and limited labeled data maximally.
As we introduced in Section~\ref{sec:MDL}, the share-private framework is the commonly used architecture for MDL.
MDCL takes the share-private framework as the backbone, where our designed loss can be easily integrated into the existing MDL methods as a plug-and-play component and trained in an end-to-end manner without introducing additional model parameters.
The inter-domain semantic alignment and the intra-domain representation learning loss are based on the contrastive learning to guild the feature extraction.
The overview of MDCL is depicted in Fig.~\ref{fig:overview}.
The remainder of this section will first provide an overview of the share-private framework, which serves as the backbone of MDCL, and then describe our method and its components in detail.

\subsection{Share-Private Framework}

The popular solution for MDL is the share-private framework (SP structure), which uses two types of feature extractors to handle both domain-shared and domain-private information.
It effectively combines shared and private feature extractors to capture domain-shared and domain-specific information in MDL. 
The shared-information across domains could improve overall performance by training a shared extractor, while domain-specific information may be lost during the joint training. 
Whereas, the private extractor can efficiently preserve domain-specific information. 
The representations from both extractors are then concatenated for inference.
Consequently, the SP model outperforms both the single shared feature extractor model (e.g., DANN \cite{DANN}) commonly used in DA, as well as private models.
Some models have unique classifiers on each domain (e.g., ASP-MTL \cite{ASP-MTL}), while others share a single classifier for all domains (e.g., MAN \cite{MAN} and CAN \cite{CAN}).
As a representative and efficient SP model, the structure of MAN is illustrated in Fig.~\ref{fig:man}.

\subsection{Inter-Domain Semantic Alignment}

The conventional idea of MDL is to utilize a shared feature extractor \cite{DANN} to align the marginal distributions of domains.
However, this approach may map instances within the same category far apart.
Therefore, it is crucial to consider the conditional distribution as well \cite{Deep_SDA,CAN}, meaning that items with identical labels should be mapped closely in the latent space, which is referred to as inter-domain semantic alignment.

Different from the previous methods for DA \cite{prototype-unsup,CLDA,ECACL}, in the scenario with limited labeled instances in MDL, obtaining reliable class prototypes is not feasible.
In this situation, we could employ a NT-Xent contrastive loss \cite{infonce} in a supervised manner \cite{supervised-CL} directly on the limited labels to proceed semantic alignment and bypass the construction of prototypes.
Pairs of items should be mapped together as long as  they belong to the same category, regardless of the domain they originate from.
As illustrated in Fig.~\ref{fig:mdcl-component}, we introduce an inter-domain semantic alignment loss on the domain-shared representation space, which can be expressed as:

\begin{equation}
    \label{euq:inter-align}
    \mathcal{L}_{\text{inter}}=\sum_{i \in I} \mathcal{L}_{i}=\sum_{i \in I} \frac{-1}{|P(i)|} \sum_{p \in P(i)} \log \frac{\exp \left(\boldsymbol{z}_i \cdot \boldsymbol{z}_p / \tau\right)}{\sum_{a \in A(i)} \exp \left(\boldsymbol{z}_i \cdot \boldsymbol{z}_a / \tau\right)}
\end{equation}

\noindent
where $i$ is the index of a sample in the augmented batch $I \equiv \left\{1\ldots 2N\right\} $, and $N$ is the batch size.
$A(i) \equiv I\setminus \left\{i\right\}$.
$P(i)$ is the index set of positives $\left\{p \in A(i): \tilde{\boldsymbol{y}}_p=\tilde{\boldsymbol{y}}_i\right\}$, and $|P(i)|$ is the cardinality.
$\boldsymbol{z}_l$ is the representation of the $l$-th item of the augmented batch extracted from the domain-shared feature extractor.
$\tau$ is a scaler temperature factor.
The original batch is chosen from a mixed labeled dataset that encompasses all domains, and then augmented to obtain $I$.
This means that instances with the same label but from different domains can be included in the same batch.
Using the semantic alignment loss, the shared feature extractor can align hidden representations of items within the same class to be close to each other.

However, solely using the proposed inter-domain semantic alignment loss may not preserve the local manifold structure and only affect on the few labeled instances.
As illustrated in Fig.~\ref{fig:intuition}, solely using inter-domain alignment might result in the unlabeled representations being mixed and losing the cluster structures.

\subsection{Intra-Domain Representative Learning}

Within each domain, unlabeled data can be utilized for intra-domain contrastive alignment.
Unlike the previous inter-domain instance alignment, we use the classifier outputs instead of intermediate representations to align items within each domain.
This approach goes beyond the intermediate representation space and ensures that the same item with different augmentations has a similar class assignment.
The contrastive alignment ensures a low density (uniformity) and form robust clusters (alignment) in the representation space \cite{alignment-uniformity}.
As a result, the domain representations are more robust to noise and augmentations.
To achieve this unsupervised alignment, we also leverage the NT-Xent contrastive loss \cite{infonce}.
As illustrated in Fig.~\ref{fig:mdcl-component}, we introduce an intra-domain contrastive loss on the domain-private output space, which can be expressed as:

\begin{equation}
    \label{euq:intra-align}
    \mathcal{L}_{\text{Intra}}=\sum_{i \in I} \mathcal{L}_i=-\sum_{i \in I} \log \frac{\exp \left(\boldsymbol{o}_i \cdot \boldsymbol{o}_{j(i)} / \tau\right)}{\sum_{a \in A(i)} \exp \left(\boldsymbol{o}_i \cdot \boldsymbol{o}_a / \tau\right)}
\end{equation}

\noindent
where $\boldsymbol{o}_l$ is the output of the classifier of the corresponding domain.

However, solely using the proposed intra-domain contrastive loss only affects the alignment in each domain.
As illustrated in Fig.~\ref{fig:intuition}, the items from different domains would still be far from each other in the representation space after solely using intra-domain alignment.

\subsection{Overall framework and Pseudocode}

Our method serves as a plug-and-play solution for MDL and can be applied to different SP models.
We provide the pseudocode for our MDAL in Algorithm \ref{alg:MDCL}.
In the pseudocode, we also take MAN \cite{MAN} as the backbone example for a better explanation.

\begin{algorithm}[t]
    \footnotesize
    \setstretch{1.13}
    \begin{algorithmic}[1]
        \Require
        labeled dataset $\mathbb{X}$; unlabeled dataset $\mathbb{U}$; hyperparameter $\lambda_d > 0$, $\lambda_{inter} > 0$, $\lambda_{intra} > 0$, $k_{inter} \in \mathbb{N}, k_{adv} \in \mathbb{N}$
        \Repeat

        \LeftComment{Inter-domain alignment}
        \For{$iter = 1$ to $k_{inter}$}
        \State $l_{inter} = 0$
        \State Sample a mini-batch $(\bm{x},\bm{y})  \sim \mathbb{X}$ \Comment{From mixed labeled dataset}
        \State $\bm{x}\prime,\bm{y}\prime$ = {Aug}$(\bm{x},\bm{y})$\Comment{Augmented batch}
        \State $\bm{z}_{s} = \cfs(\bm{x})$; $\bm{z}_{s}^\prime = \cfs(\bm{x}\prime)$  \Comment{Shared feature vector}
        \State $l_{inter} = \lambda_{inter} \cdot L_{inter}(\bm{z}_{s}, \bm{z}_{s}^\prime; \bm{y}, \bm{y}\prime)$ \Comment{\textbf{Inter-domain loss (Equ. \ref{euq:inter-align})}}
        \State Update $\cfs$ parameters using $\nabla l_{inter}$
        \EndFor
        \LeftComment{Discriminator training}
        \For{$iter = 1$ to $k_{adv}$}
        \State $l_\cd = 0$
        \ForAll{$d \in \Delta$}\Comment{For all $N$ domains}
        \State Sample a mini-batch $\bm{x} \sim \mathbb{U}_{d}$
        \State $\bm{z}_s = \cfs(\bm{x})$
        \State $l_\cd \pluseq L_\cd(\cd(\bm{z}_s); d)$ \Comment{Accumulate $\cd$ loss}
        \EndFor
        \State Update $\cd$ parameters using $\nabla l_\cd$
        \EndFor

        \LeftComment{Main iteration}
        \State $loss = 0$
        \ForAll{$d \in \Delta_L$}\Comment{For all labeled domains}
        \State Sample a mini-batch $(\bm{x},\bm{y})  \sim \mathbb{X}_{d}$
        \State $\bm{z}_s = \cfs(\bm{x})$
        \State $\bm{z}_d = \cfd(\bm{x})$ \Comment{Domain feature vector}
        \State $loss \pluseq L_\cc(\cc(\bm{z}_s, \bm{z}_d); \bm{y})$ \Comment{Compute $\cc$ loss}
        \EndFor
        \ForAll{$d \in \Delta$}\Comment{For all $N$ domains}
        \State Sample a mini-batch $\bm{x} \sim \mathbb{U}_{d}$
        \LeftComment{Discriminate}
        \State $\bm{z}_s = \cfs(\bm{x})$ 
        \State $loss \pluseq \lambda_d\cdot L_{\cfs}^\cd(\cd(\bm{z}_s); d)$ \Comment{Domain loss of $\cfs$}
        \LeftComment{Intra-domain alignment}
        \State $\bm{x}\prime$ = {Aug}$(\bm{x})$\Comment{Augmented batch}
        \State $\bm{o} = \cc(\cfs(\bm{x}, \cfd(\bm{x})); \bm{o}\prime = \cc(\cfs(\bm{x}\prime, \cfd(\bm{x}\prime));$    \Comment{Output of the classifier}
        \State $loss \pluseq \lambda_{intra} \cdot L_{intra}(\bm{o}, \bm{o}\prime)$ \Comment{\textbf{Intra-domain loss (Equ. \ref{euq:intra-align})}}

        \EndFor
        \State Update $\cfs$, $\cfd$, $\cc$ parameters using $\nabla loss$
        \Until{convergence}
    \end{algorithmic}
    \caption{Multi-Domain Contrastive Learning.}
    \label{alg:MDCL}
\end{algorithm}

\begin{table*}[htbp]
    \caption{The hyperparameters used for MDCL.}
    \label{tab:hyperparameters}
    \centering
    \scalebox{1}{
        \begin{tabular}{cccccccccc}
            \toprule
            Datasets    & Optimizer & \makecell[c]{Learning                                              \\Rate} & \makecell[c]{Learning  \\Rate Decay} & \makecell[c]{Batch \\Size}  & \makecell[c]{Weight \\Decay} & \makecell[c]{Early \\Stopping}   & \makecell[c]{Inter-Domain\\$\lambda$ \& $\tau$} & \makecell[c]{Intra-Domain\\$\lambda$ \& $\tau$}\\
            \midrule
            Amazon      & Adam      & 3e-4                  & False & 8 & 0.05  & 20 & 1/0.1    & 1/0.01 \\
            MNIST-USPS  & Adam      & 3e-3                  & 0.33  & 8 & 0.001 & 30 & 0.1/0.1  & 1/0.1  \\
            Office-Home & Adam      & 1e-2                  & 0.33  & 8 & 0.001 & 15 & 1/0.01   & 1/0.01 \\
            FDUMTL      & Adam      & 3e-4                  & 0.1   & 8 & 0.001 & 30 & 0.1/0.01 & 1/0.01 \\
            PACS        & SGD       & 1e-3                  & 0.1   & 8 & 0.001 & 15 & 0.1/1    & 1/0.1  \\
            \bottomrule
        \end{tabular}
    }
\end{table*}

\section{Experiments}

\subsection{Research Questions}

We evaluate our method on the following research questions:
\begin{enumerate}
    \item As a plug-and-play method, can MDCL enhance the performance of various models under the SP structure with the limited number of labeled instances (around 5\%-20\%) or extremely few labeled instances (around 1\%)? (Section \ref{sec:few-label-performance})
    \item How does each component of MDCL affect the performance? (Section \ref{sec:ablation})
    \item Given a further labeling budget, MDAL could be utilized. Can MDCL improve the entire MDAL process with a relatively large number of unlabeled instances (5\%-50\%)? (Section \ref{sec:mdal})
\end{enumerate}

\subsection{Experimental Setup}

\subsubsection{Dataset}
We evaluate our proposed MDCL method on several popular multi-domain textual and image datasets, namely Amazon \cite{mSDA}, MNIST-USPS \cite{ADDA}, Office-Home \cite{office-home}, FDUMTL \cite{ASP-MTL}, and PACS \cite{PACS}.
The Amazon dataset consists of four textual domains, each containing two categories, with instances encoded to a vector representation of length 5000.
MNIST-USPS comprises two image domains with ten categories each, and instances are encoded to a vector representation of length 256.
The Office-Home dataset contains four image domains with 65 categories each, with instances encoded to a vector representation of length 2048.
The FDUMTL dataset comprises sixteen textual domains, and we use the first four of them in our experiment, each containing two categories, with raw texts utilizing word2vec embedding.
Finally, the PACS dataset consists of four image domains with seven categories each, and raw images are preprocessed to tensor with shape (224,224,3).

\subsubsection{Models}

Our primary focus is evaluating the performance of MDCL on a single model.
Therefore, we selected the most renowned and widely recognized models in MDL as baselines.
In the majority of the experiments, MAN \cite{MAN} is utilized as the backbone of MDCL due to its simple structure and wide acceptance in the literature.
ASP-MTL \cite{ASP-MTL}, as the most classic MDL model, is also used to verify the generation ability of MDCL.
Compared to MAN, ASP-MTL has specific classifiers for each domain, while MAN has a shared classifier for all domains.
While there are newer models available, such as CAN \cite{CAN}, the existing literature \cite{mdal} suggests that these models do not outperform MAN.
Some other recently proposed models in the MDL are designed for particular tasks \cite{MDNMT,mdl-medical-image}, which are beyond the scope of our paper.

The balancing parameter $\lambda_d$ for $l_\cd$ is set to 0.05 for all the datasets.
For Amazon, MNIST-USPS and Office-Home datasets, we use one fully connected layer with a sigmoid activation function as the $\cfs$ and $\cfd$ feature extractor, both with the same output size of 64.
For the FDUMTL dataset, we use a CNN as the feature extractor that takes 100d word embeddings as input from the sequence, and the output representation length is 128.
A single convolution layer with 200 kernels of sizes 3, 4, and 5 is used.
For the PACS dataset, we use a pre-trained Resnet-18 \cite{ResNet18} as the feature extractor with an output size of 64.
For all the datasets, the classifier and domain discriminator consist of a single fully connected layer.
Table~\ref{tab:hyperparameters} provides all the hyper-parameters for optimizations in training details for each dataset.

\subsubsection{MDCL Implementation Details}

For MDCL, the inter-domain and intra-domain contrastive learning requires a balancing ($\lambda$) and a temperature ($\tau$) hyperparameter, which are listed in Table~\ref{tab:hyperparameters}.
For the FDUMTL dataset, multiple outputs from dropout layers are utilized to obtain batch augmentation.
For the rest datasets, we apply batch augmentation by using Gaussian noise with a standard deviation of 0.01.
All the experiments were conducted five times, and the average performance and standard deviation were calculated to ensure the reliability of the results.


\subsection{RQ1: Performance with insufficient labels}
\label{sec:few-label-performance}

We assess the efficacy of our MDCL method with a limited number of labeled instances.
First, two scenarios are included: 1) when the number of labeled instances is moderately limited (around 5\%-20\%), and 2) when the number of labeled instances is extremely low (1\%).
Both scenarios are evaluated on the MAN model.
Then, ASPMTL is used to verify the generation ability of MDCL over SP models in the limited labeled case (around 5\%).

\begin{figure*}[ht]
    \centering
    \hspace{-0.03\linewidth}
    \subfigure[Amazon]{
        \begin{minipage}[b]{0.33\linewidth}
            \includegraphics[width=1\textwidth]{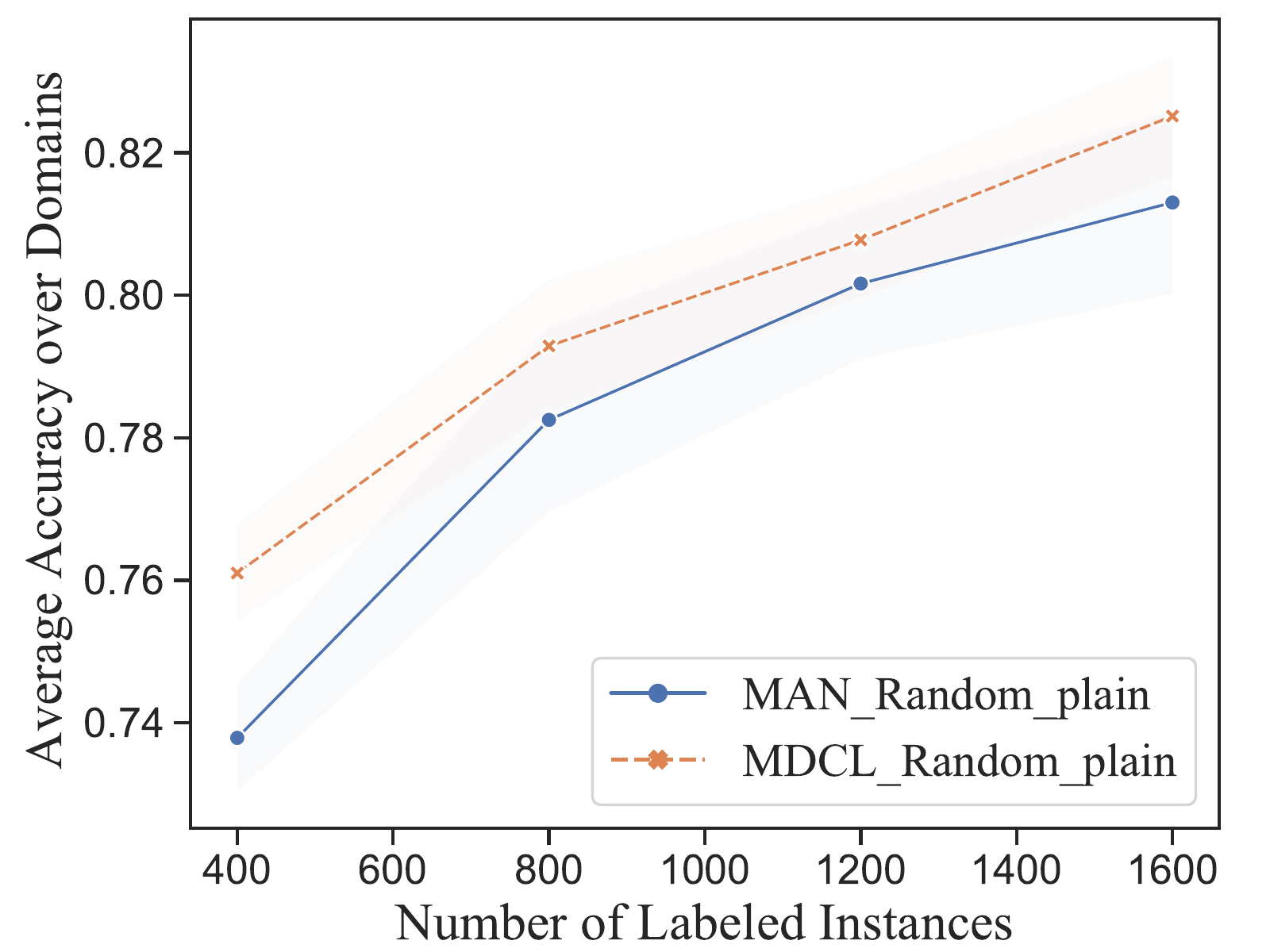}
        \end{minipage}
        \label{fig:few_amazon}
    }
    \hspace{-0.03\linewidth}
    \subfigure[MNIST-USPS]{
        \begin{minipage}[b]{0.33\linewidth}
            \includegraphics[width=1\textwidth]{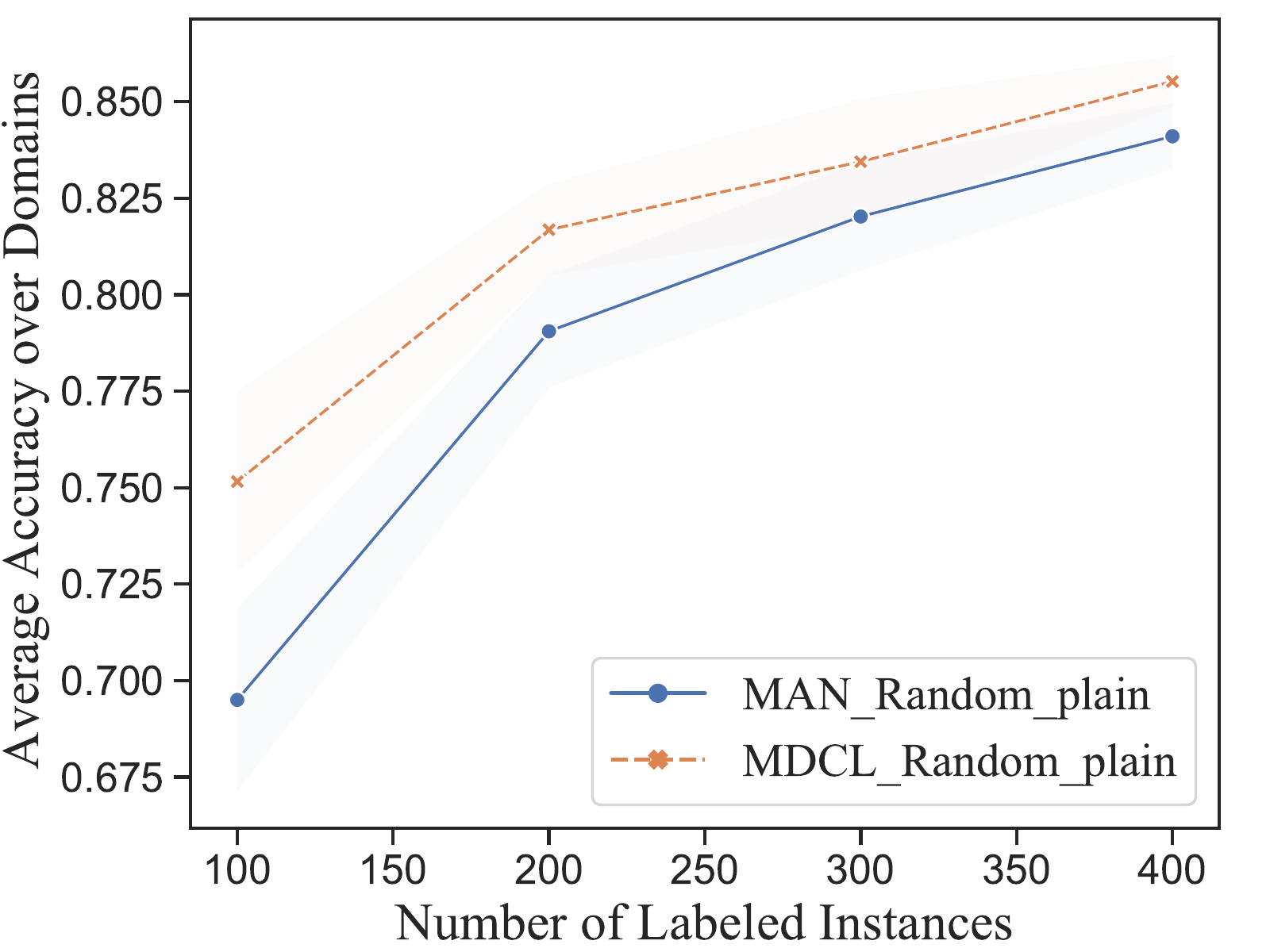}
        \end{minipage}
    }
    \hspace{-0.03\linewidth}
    \subfigure[Office-Home]{
        \begin{minipage}[b]{0.33\linewidth}
            \includegraphics[width=1\textwidth]{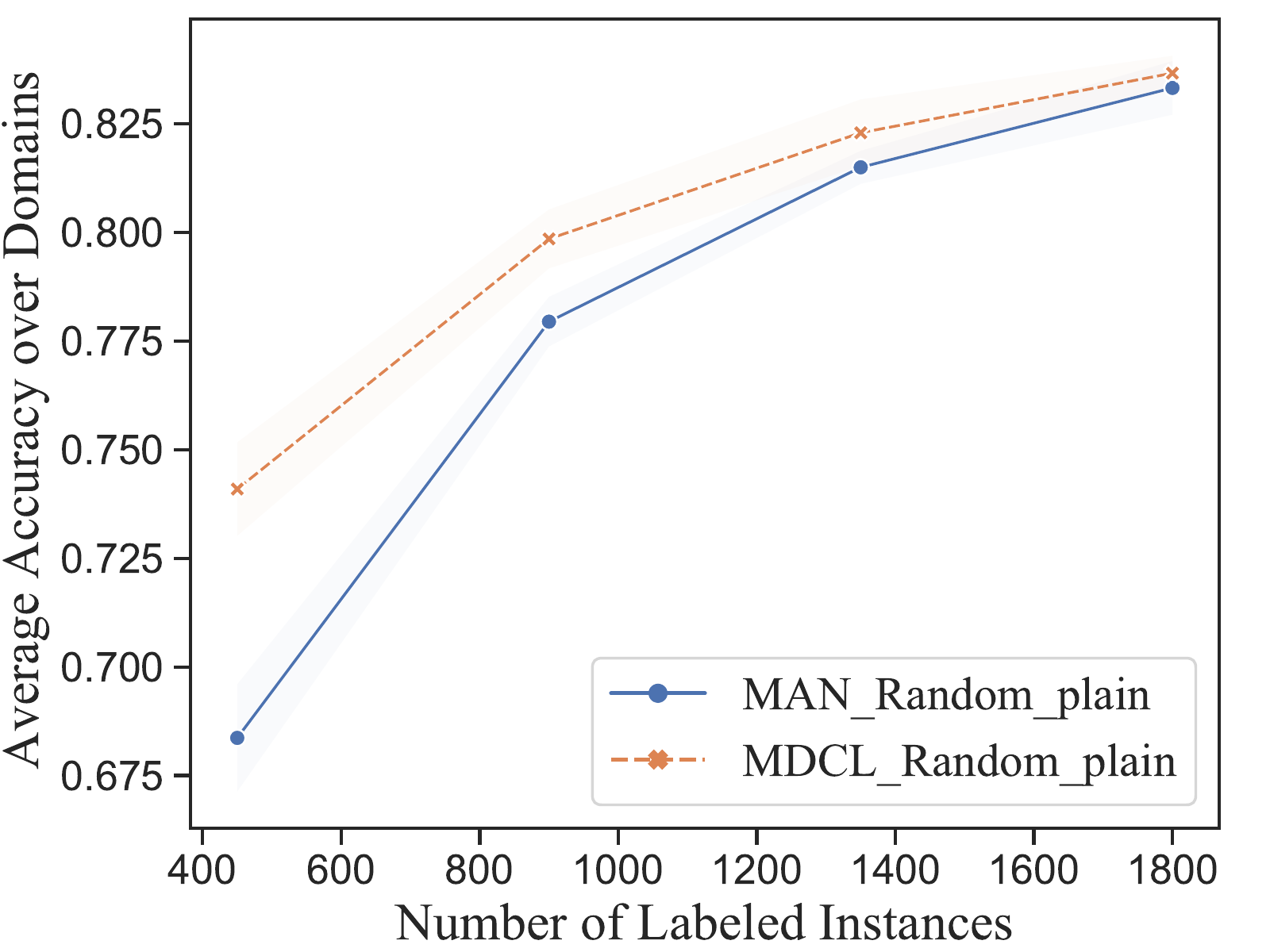}
        \end{minipage}
        \label{fig:shallow-amazon}
    }
    \\
    \hspace{-0.03\linewidth}
    \subfigure[FDUMTL (4 domains)]{
        \begin{minipage}[b]{0.33\linewidth}
            \includegraphics[width=1\textwidth]{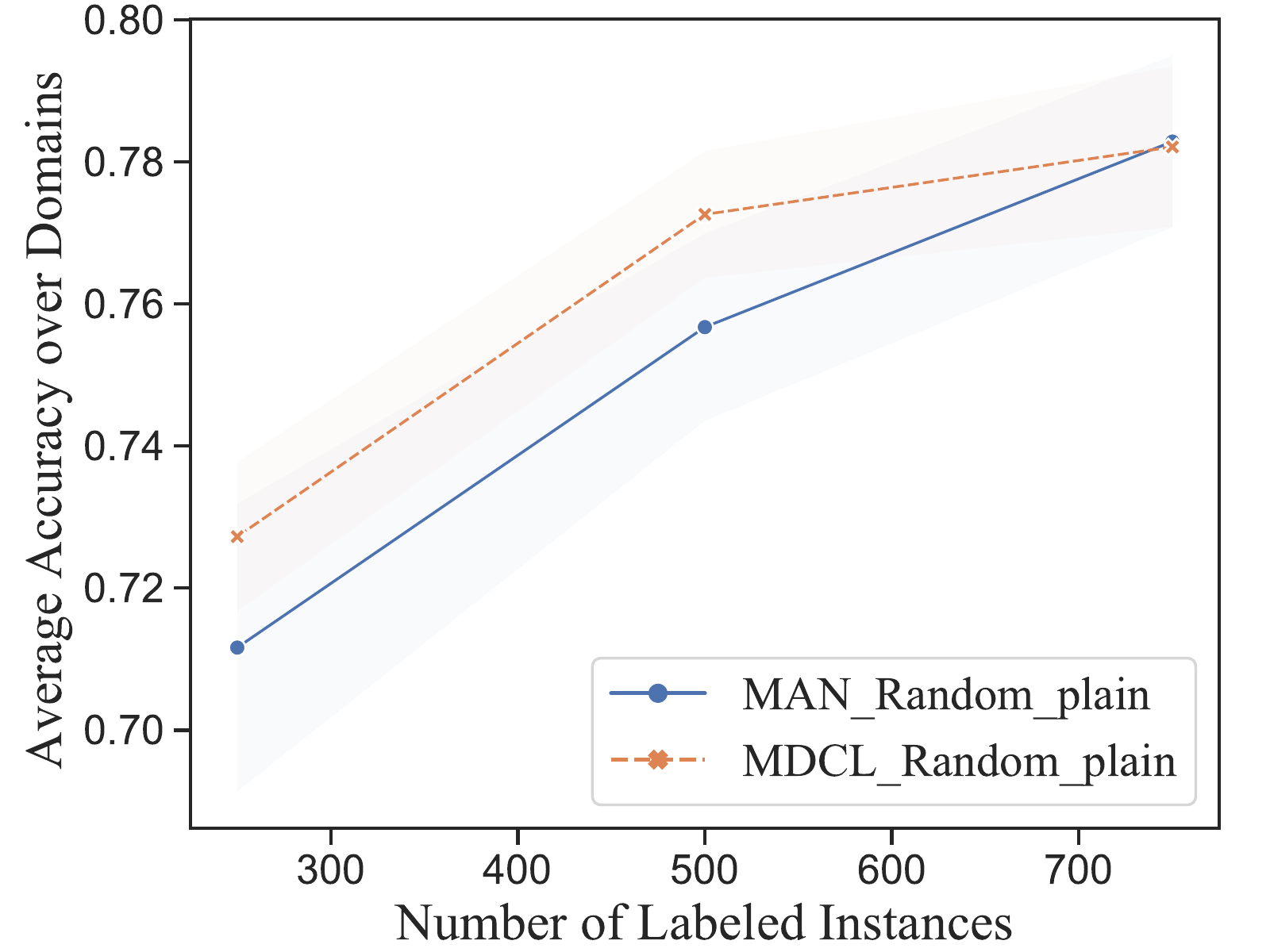}
        \end{minipage}
        \label{fig:deep-fdumtl4}
    }
    \hspace{-0.03\linewidth}
    \subfigure[FDUMTL]{
        \begin{minipage}[b]{0.33\linewidth}
            \includegraphics[width=1\textwidth]{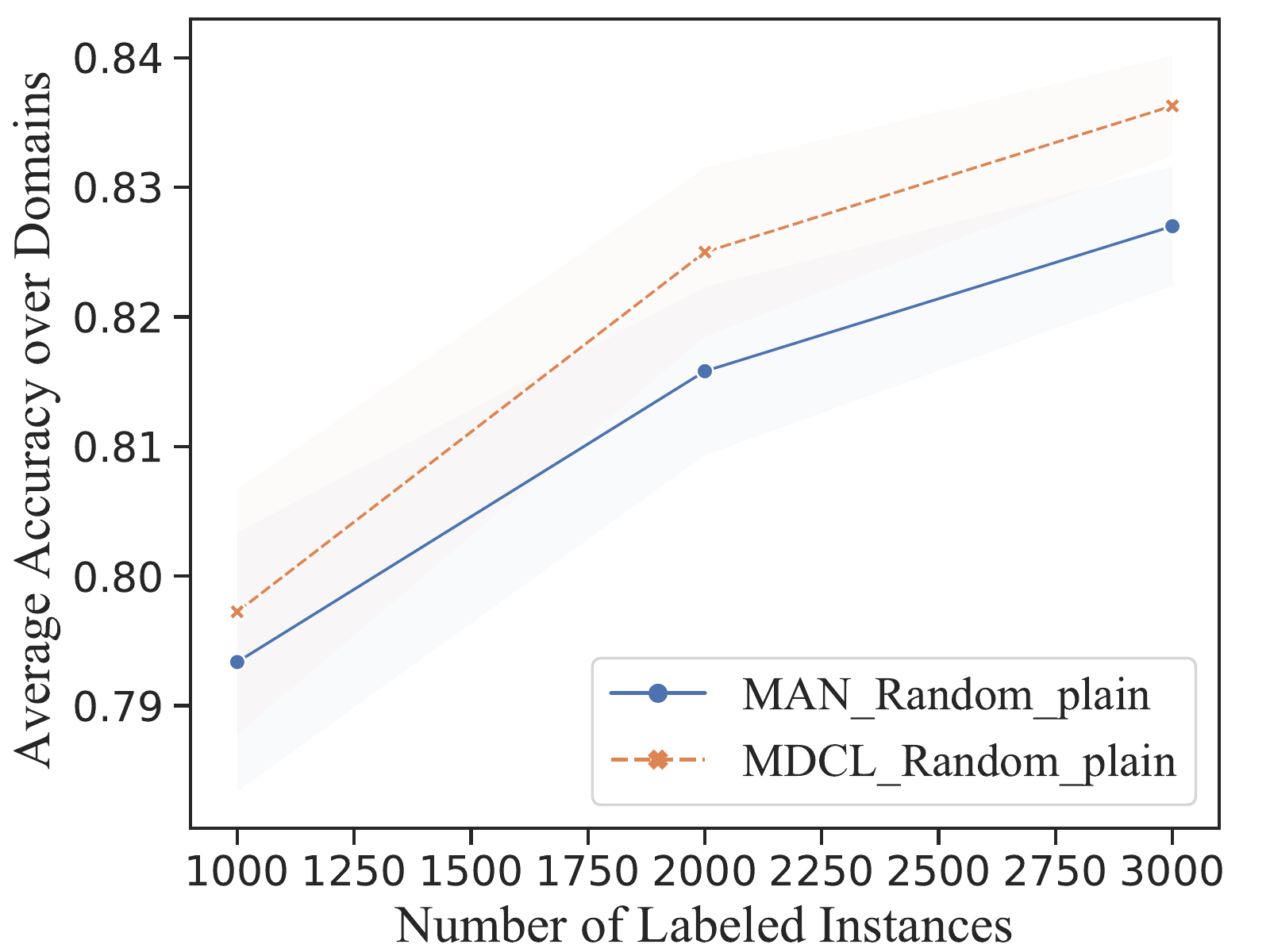}
        \end{minipage}
        \label{fig:deep-fdumtl}
    }
    \hspace{-0.03\linewidth}
    \subfigure[PACS]{
        \begin{minipage}[b]{0.33\linewidth}
            \includegraphics[width=1\textwidth]{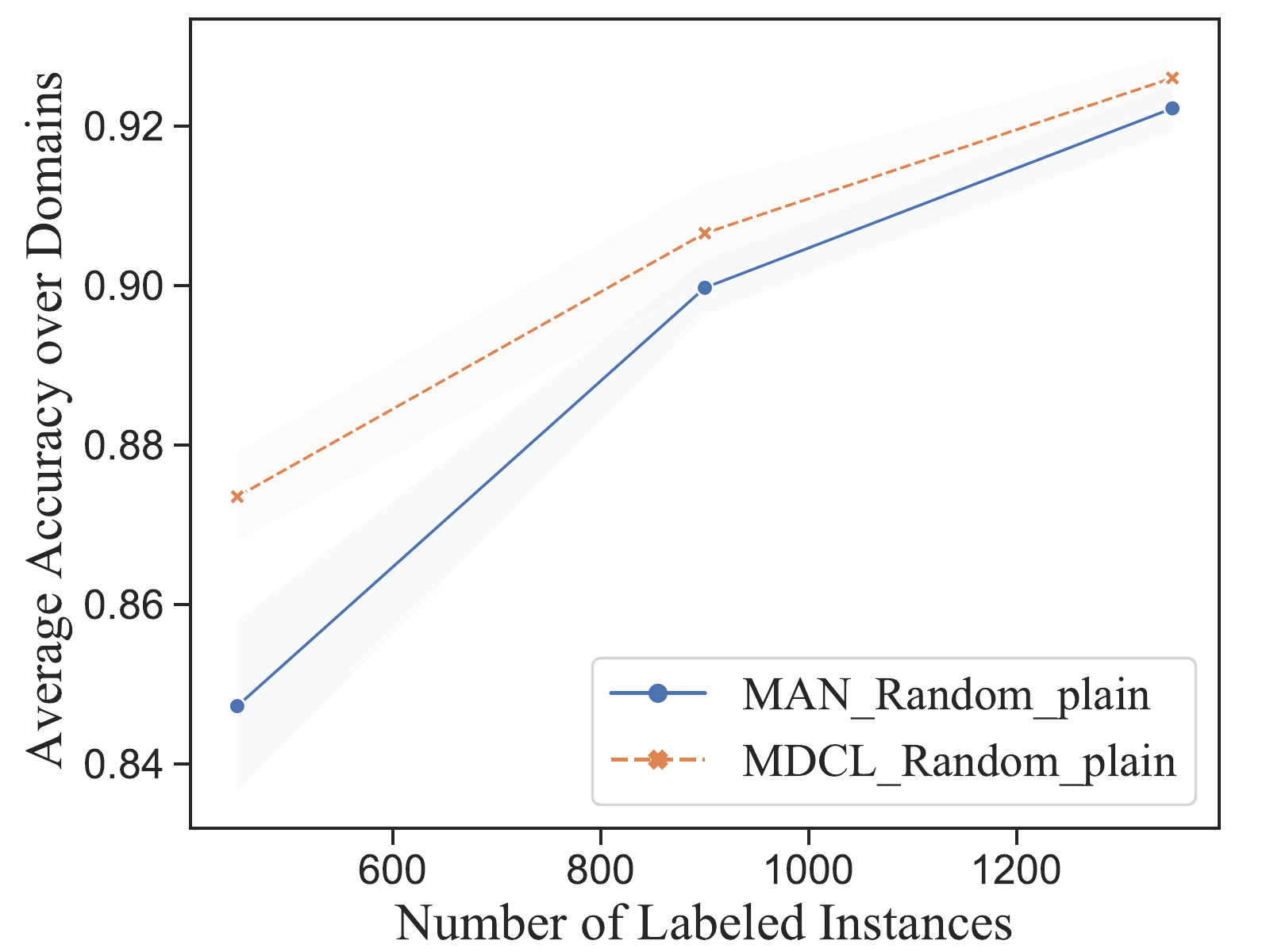}
        \end{minipage}
        \label{fig:deep-digits}
    }
    \caption{The results of MDCL with different number of labeled instances on different datasets.}
    \label{fig:few}
\end{figure*}

\paragraph{Moderately insufficient labels}
We evaluate our method on datasets Amazon, MNIST-USPS, and Office-Home with 5\% to 20\% labeled instances, and on datasets FDUMTL and PACS with 5\% to 15\% labeled instances.
The performance results are illustrated as learning curves in Figure~\ref{fig:few}.
On all the datasets, MDCL brings obvious improvement on solely using MAN.
The improvement is usually more significant when the number of labeled instances is low and decreases with more labeled instances.

\begin{table}
    \centering
    \caption{MDCL on only 1\% labeled instances. Average performance in 5 runs with the standard deviation in parentheses.}
    \label{table:1percent}
    \scalebox{0.9}{
        \begin{tabular}{c|c|c|c|c|c}
            \toprule
            Method & Amazon  & MNIST-USPS & Office-Home & FDUMTL & PACS \\
            \midrule
            MAN    & \makecell[c]{0.6374 \\ (0.0180)}& \makecell[c]{0.3886 \\ (0.0831)} & \makecell[c]{0.347 \\ (0.0248)}& \makecell[c]{0.5204\\(0.0163)} & \makecell[c]{0.6723 \\ (0.0418)} \\
            \makecell[c]{MDCL\\(+MAN)} & \makecell[c]{0.6631 \\ (0.0292)}& \makecell[c]{0.4750\\(0.1205)} & \makecell[c]{0.421 \\ (0.0315)}& \makecell[c]{0.5592\\(0.0559)} & \makecell[c]{0.6911 \\ (0.0290)} \\
            \bottomrule
        \end{tabular}
    }
\end{table}

\paragraph{Extremely insufficient labels}

We evaluated the performance of MDCL with an extremely small number of labeled instances on Amazon, FDUMTL, and PACS datasets.
Only used 1\% labeled instances are used to train the model.
The results are presented in Table~\ref{table:1percent}.
The improvement is also obvious on all the datasets.
The improvement is more significant on the Office-Home dataset, which has more categories than the other two datasets.

\begin{table}
    \caption{MDCL with ASPMTL on 1\% \& 5\% labeled instances. Average performance in 5 runs with the standard deviation in parentheses.}
    \centering
    \scalebox{1}{
        \begin{tabular}{cccccccccccc}
            \toprule
            \multirow{2}[3]{*}{Method} & \multicolumn{2}{c}{Amazon} & \multicolumn{2}{c}{Office-Home}             \\
            \cmidrule(l){2-3} \cmidrule(l){4-5}
                                       & 1\%                        & 5\%                             & 1\% & 5\% \\
            \midrule
            ASPMTL                     & \makecell[c]{0.5969                                                      \\ (0.0147)}    & \makecell[c]{0.6874 \\ (0.0128)}  &  \makecell[c]{0.1665 \\ (0.0127)}  &    \makecell[c]{ 0.5816 \\ (0.016)}   \\
            \makecell[c]{MDCL                                                                                     \\(+ASPMTL)} & \makecell[c]{0.617 \\ (0.0164)} & \makecell[c]{0.7224 \\ (0.0221)}   & \makecell[c]{0.2348 \\ (0.0242)} & \makecell[c]{0.6141 \\  (0.0125)} \\
            \bottomrule
        \end{tabular}
    }
    \label{table:aspmtl}
\end{table}

\paragraph{Compatibility for share-private framework}
As a plug-and-play method, MDCL should be compatible with other SP models.
We compare the performance of our MDCL method with another popular model, ASPMTL \cite{ASP-MTL}, on the Amazon and Office-Home datasets with 5\% labeled instances to evaluate its ability to be combined with other models.
The results are shown in Table~\ref{table:aspmtl}.
MDCL outperforms ASPMTL on both datasets, which validates the generalization ability of MDCL on different SP models.

\subsection{RQ2: Effectiveness of components}
\label{sec:ablation}

We explore the effectiveness of the two components of MDCL, i.e., the inter-domain and the intra-domain contrastive loss, by conducting ablation studies on the PACS and MNIST-USPS datasets with 5\% labeled instances.
The results are shown in Table~\ref{table:ablation}.
The inter-domain contrastive proves more effective than the intra-domain contrast on PACS, but the opposite is observed for MNIST-USPS.
This discrepancy may be attributed to the fact that, on PACS, the pre-trained feature extractor already provides a reliable domain-invariant feature representation, making semantic information more critical.
In contrast, on MNIST-USPS, the intra-domain contrastive loss is more effective, as it assists the feature extractor in learning features in each domain.

\begin{table}
    \centering
    \caption{Ablation study on PACS and MNIST-USPS datasets.}
    \label{table:ablation}
    \scalebox{1}{
        \begin{tabular}{c|c|c}
            \toprule
            Method                & MNIST-USPS & PACS   \\
            \midrule
            MAN                   & 0.7007     & 0.8478 \\
            MDCL \textit{(Inter)} & 0.7024     & 0.8717 \\
            MDCL \textit{(Intra)} & 0.7505     & 0.852  \\
            MDCL                  & 0.7575     & 0.8748 \\
            \bottomrule
        \end{tabular}
    }
\end{table}

\begin{figure*}[htbp]
    \centering
      \hspace{-0.03\textwidth}
    \subfigure[Amazon]{
        \begin{minipage}[b]{0.33\textwidth}
            \includegraphics[width=1\textwidth]{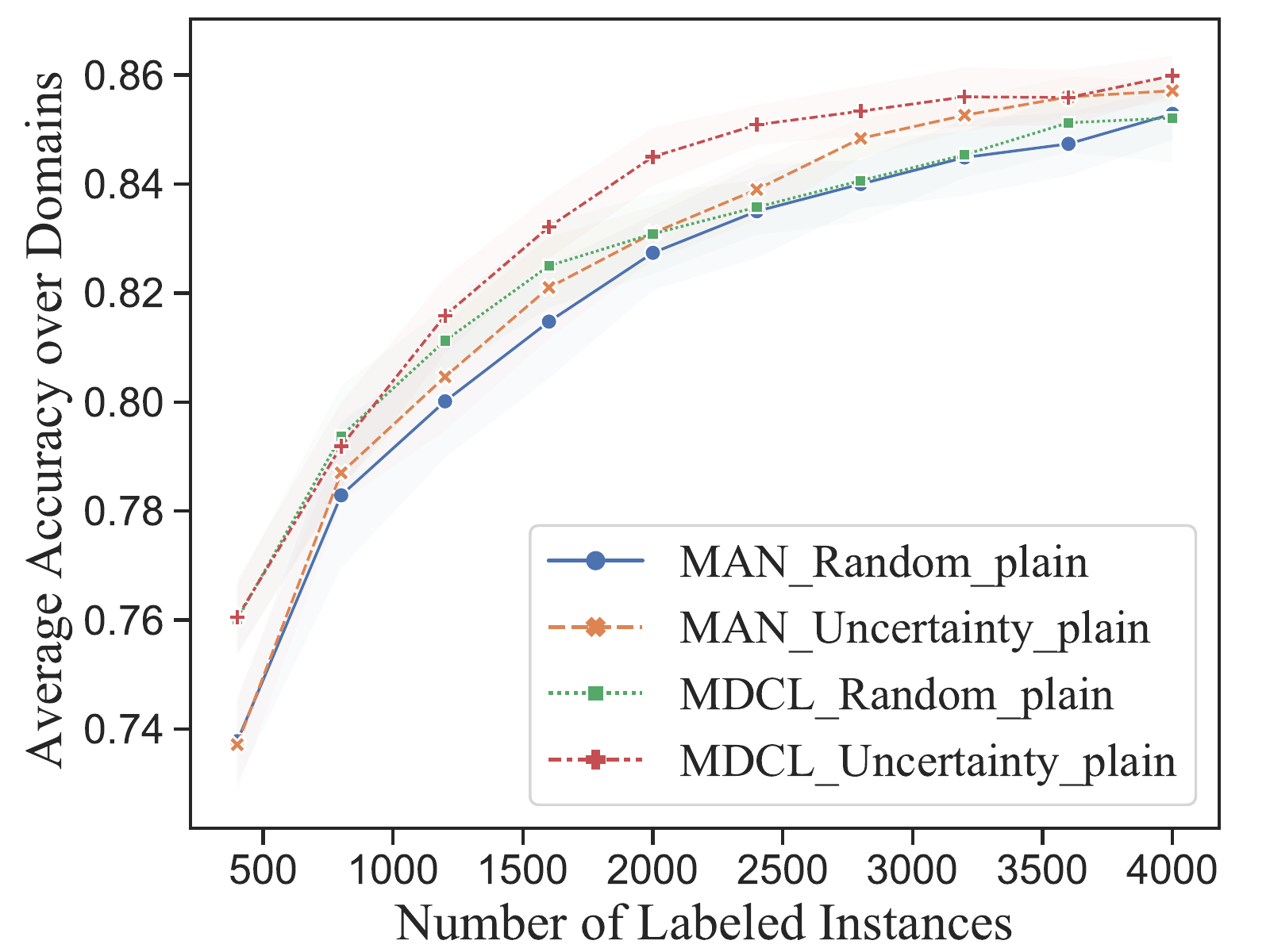}
        \end{minipage}
        \label{fig:mdal-office-home}
    }
      \hspace{-0.03\textwidth}
    \subfigure[Office-Home]{
        \begin{minipage}[b]{0.33\textwidth}
            \includegraphics[width=1\textwidth]{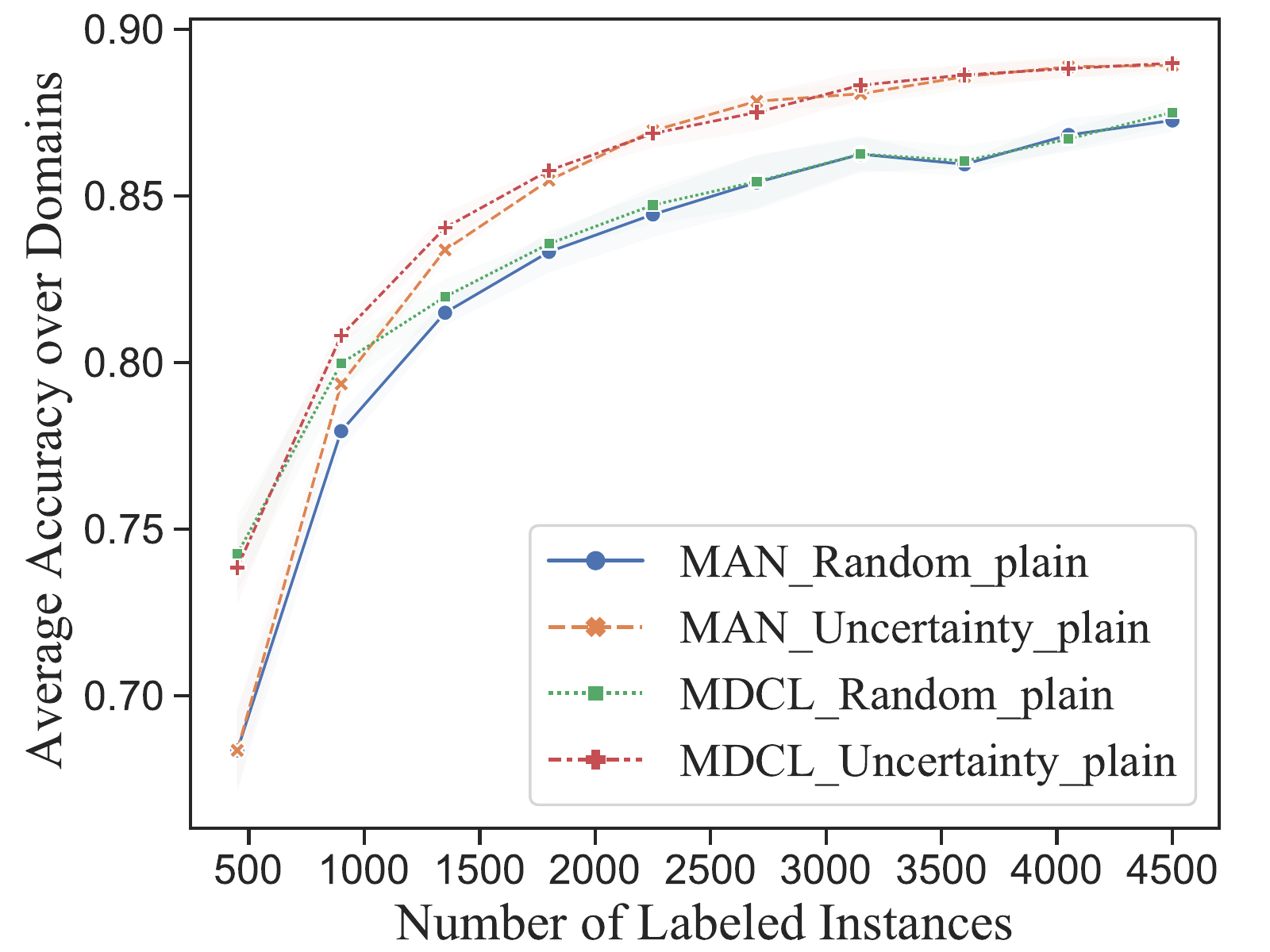}
        \end{minipage}
    }
      \hspace{-0.03\textwidth}
    \subfigure[MNIST-USPS]{
        \begin{minipage}[b]{0.33\textwidth}
            \includegraphics[width=1\textwidth]{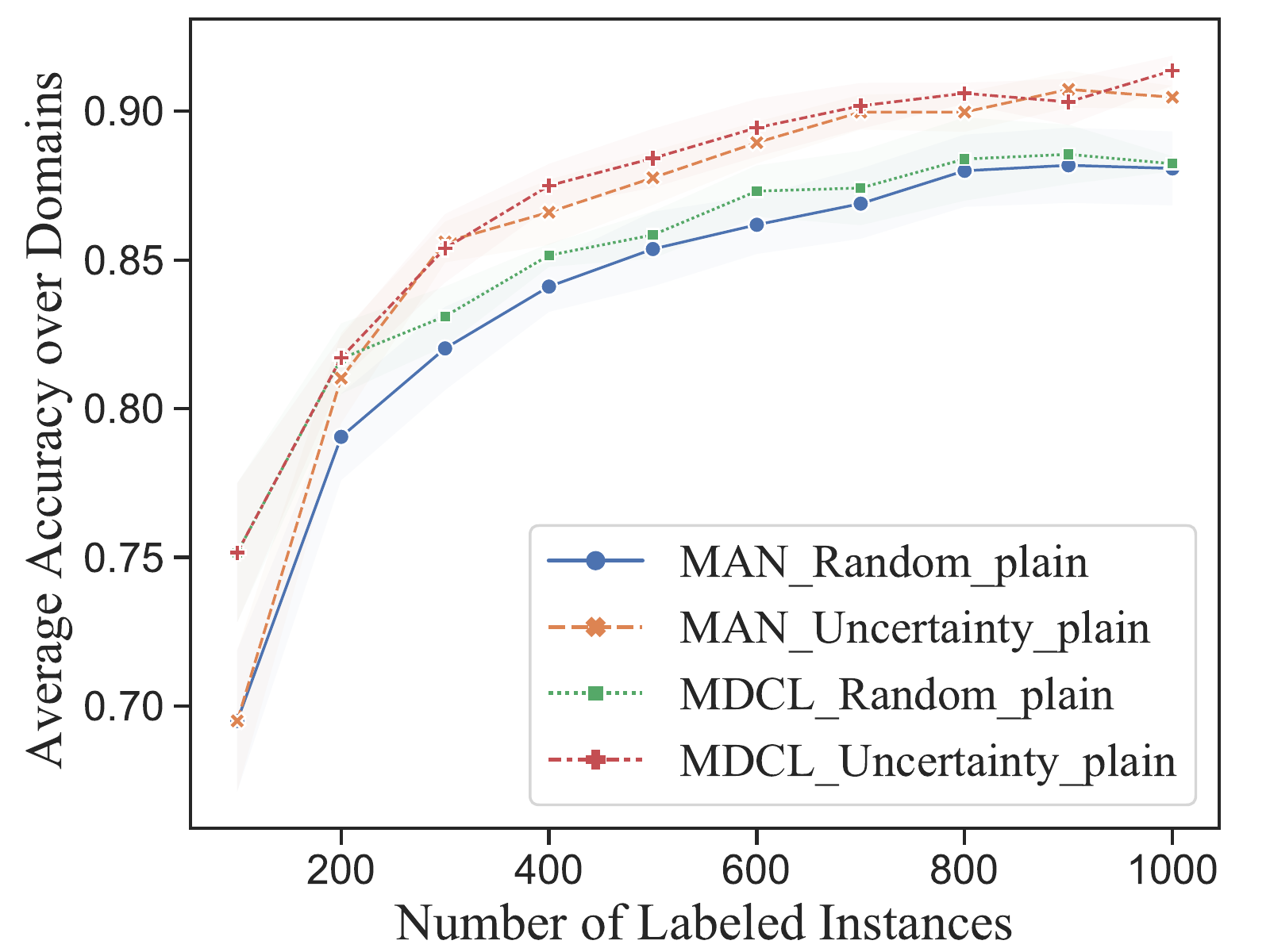}
        \end{minipage}
    }
    \caption{The results of MDCL combined with the Uncertainty strategy in a MDAL setting.}
    \label{fig:mdal}
\end{figure*}

\subsection{RQ3: Ability to integrate with MDAL}
\label{sec:mdal}

In section~\ref{sec:few-label-performance}, the improvement decreases with more labeled instances.
In real applications, a method that only performs well with fewer labeled instances but poorly with more labeled instances is not desirable.
Thus, the proposed method should be evaluated with more labeled instances.
We consider a setting where there is a larger budget for collecting labeled instances, which allows for more labeled instances to be included in the learning process.
Specifically, we can use the MDAL method to iteratively collect informative labeled instances, with MDCL being used in each iteration.

We conduct experiments with 5\% to 50\% labeled instances on the Amazon, MNIST-USPS, and Office-Home datasets using the simplest but effective active learning (AL) query strategy, Best-vs-Second-Best \cite{BvSB} (BvSB).
The learning curves on all the datasets are shown in Figure~\ref{fig:mdal}.
Moreover, the area-under-learning-curves (AULC) results of all three datasets are presented in Table~\ref{table:mdal}.

\begin{table}
    \centering
    \caption{AULC of MDCL. Average performance in 5 runs with the standard deviation in parentheses.}
    \label{table:mdal}
    \scalebox{1}{
        \begin{tabular}{c|c|c|c}
            \toprule
            Method & Amazon & \makecell[c]{MNIST \\ -USPS}  & \makecell[c]{Office \\ -Home} \\
            \midrule
            \makecell[c]{MAN                     \\ \textit{+Random}} & \makecell[c]{82.09 \\(0.46)} & \makecell[c]{84.29 \\(1.00)}    & \makecell[c]{83.28 \\(0.23)}  \\
            \makecell[c]{MDCL                    \\ \textit{+Random}} & \makecell[c]{82.67 \\(0.37)} & \makecell[c]{85.46 \\(0.68)} & \makecell[c]{83.96 \\(0.15)}  \\
            \midrule
            \makecell[c]{MAN                     \\ \textit{+BvSB}}   & \makecell[c]{82.63 \\(0.31)} & \makecell[c]{86.74 \\(0.50)} & \makecell[c]{85.25 \\(0.09)}  \\
            \makecell[c]{MDCL                    \\ \textit{+BvSB}}   & \makecell[c]{83.46 \\(0.18)} & \makecell[c]{87.43 \\(0.37)} & \makecell[c]{85.81 \\(0.11)}  \\
            \bottomrule
        \end{tabular}
    }
\end{table}

From the results, we can see that the proposed MDCL method performs well on fewer labeled instances and performs competitively to MAN with more labeled instances.
Integrated with MDAL, MDCL further improves performance.
With BvSB, MDCL starts from a better initialization and could provide more reliable selections for active learning.
In Figure~\ref{fig:mdal}, MDCL dominates MAN in the whole MDAL learning process.
In Table~\ref{table:mdal}, MDCL obtains higher AULC scores on all the datasets in both passive and active learning settings.

\section{Conclusion}

In conclusion, we introduce a novel multi-domain contrastive learning (MDCL) approach for multi-domain learning.
The primary objective of MDCL is to build neural network models on insufficient labeled instances from multiple domains.
MDCL comprises two components: a supervised contrastive loss for inter-domain semantic alignment and an unsupervised contrastive loss for intra-domain representation learning.
MDCL is readily compatible with many SP models, requiring no additional model parameters and allowing for end-to-end training.
Experimental results across five textual and image multi-domain datasets demonstrate that MDCL brings noticeable improvement over various SP models.
Additionally, given a labeling budget, MDCL can be further employed in multi-domain active learning to enhance the performance of the entire learning process.

\ack 
This work was supported in part by the National Key Research and Development Program of China under Grant 2022YFA1004102 and in part by the National Natural Science Foundation of China under Grant 62250710682.

\bibliography{mdcl}
\end{document}